\documentclass{article}

\PassOptionsToPackage{numbers, compress}{natbib}
\usepackage[preprint]{neurips_2026}

\usepackage[utf8]{inputenc} 
\usepackage[T1]{fontenc}    
\usepackage{hyperref}       
\usepackage{url}            
\usepackage{booktabs}       
\usepackage{amsfonts}       
\usepackage{nicefrac}       
\usepackage{microtype}      
\usepackage{xcolor}         
\usepackage{amsmath}
\usepackage{graphicx}
\usepackage{booktabs}
\usepackage{amssymb}
\usepackage{multirow}
\usepackage{wrapfig}
\usepackage{caption}
\usepackage{enumitem}
\usepackage{xcolor}

\title{Image2Sim: Scaling Embodied Navigation via Generative Neural Simulator}

\author{
  Zihan Wang\textsuperscript{1}, \quad Seungjun Lee\textsuperscript{1}, \quad 
  Yinghao Xu\textsuperscript{2}, \quad
  Gim Hee Lee\textsuperscript{1} \\
\textsuperscript{1}National University of Singapore, \textsuperscript{2}HKUST \\
    {\tt
    zihan.wang@u.nus.edu, gimhee.lee@nus.edu.sg
    }
}

\begin{document}

\maketitle

\begin{abstract}
Embodied navigation aims to build agents that interpret multimodal goals, reason in 3D space, and reach target destinations reliably in the real world. However, progress remains constrained by the lack of scalable, high-fidelity, and physically grounded interactive environments. Although real-world scanned datasets offer visual realism, they are limited by scale. In contrast, synthetic simulators scale more easily but often exhibit large sim-to-real gaps. We introduce Image2Sim, a real-time neural simulation framework that constructs high-quality interactive environments from posed RGB-D image sequences. The central idea is to decouple \emph{3D spatial anchoring} from \emph{photorealistic observation synthesis}. For scene construction, Image2Sim uses a feed-forward feature Gaussian model that lifts posed RGB-D observations into a 3D feature-Gaussian representation in a single pass. For rendering, we propose a Geometry-Aware One-Step Pixel Flow model that transforms sparse and noisy Gaussian projections into high-quality panoramic RGB-D observations. Image2Sim also serves as a fully automated embodied data engine that generates high-fidelity observations, executable actions, and diverse navigation instructions at scale. It converts large collections of videos and images into near 20K interactive scenes and synthesizes more than 10 million navigation training samples. Navigation models trained entirely in these neural environments achieve strong improvements on major benchmarks and transfer effectively to real-world zero-shot settings. These results suggest that scalable neural simulation can serve as a practical training substrate for embodied navigation at scale. \textbf{Project page:} 
\href{https://github.com/MrZihan/Image2Sim}
{\textcolor{blue}{\texttt{github.com/MrZihan/Image2Sim}}}
\end{abstract}

\section{Introduction}
Large language models~\cite{achiam2023gpt,team2023gemini,touvron2023llama}, vision foundation models~\cite{kirillov2023segment,simeoni2025dinov3,wang2025vggt}, and visual generative models~\cite{rombach2022high,peebles2023scalable,assran2025v} suggest that scaling data can lead to substantial gains in capability. 
However, similar scaling in embodied navigation~\cite{anderson2018vision,krantz2020beyond,rxr,qi2020reverie, chaplot2020object} has not delivered comparable progress in robust real-world generalization, largely due to the lack of suitable data. 
A key limitation lies in the availability of scalable sources of high-fidelity, physically grounded, and interactive 3D environments.

Existing data sources expose a fundamental tradeoff between visual fidelity and scalability. 
Training data is typically drawn from two primary sources: real-world 3D scans~\cite{chang2017matterport3d,ramakrishnan2habitat,xia2018gibson,savva2019habitat} and synthetic procedural environments~\cite{deitke2022️,khanna2024habitat,kolve2017ai2}. 
Real-world scans provide strong visual fidelity, but are expensive to acquire and difficult to scale. 
Procedural environments improve scalability, but often introduce substantial sim-to-real gaps due to unrealistic assets, layouts, and rendering statistics. 
Recent generative video models~\cite{blattmann2023stable, bruce2024genie, bar2025navigation} offer another potential source of realistic visual data. 
However, they struggle to maintain the rigid-body consistency and explicit collision structure required for reliable closed-loop interaction. 
These limitations suggest that scalable embodied data requires both real-world visual fidelity and explicit 3D physical grounding.

\begin{figure*}
\noindent\begin{minipage}[h!]{1\columnwidth}%
\begin{center}
\includegraphics[width=0.95\columnwidth]{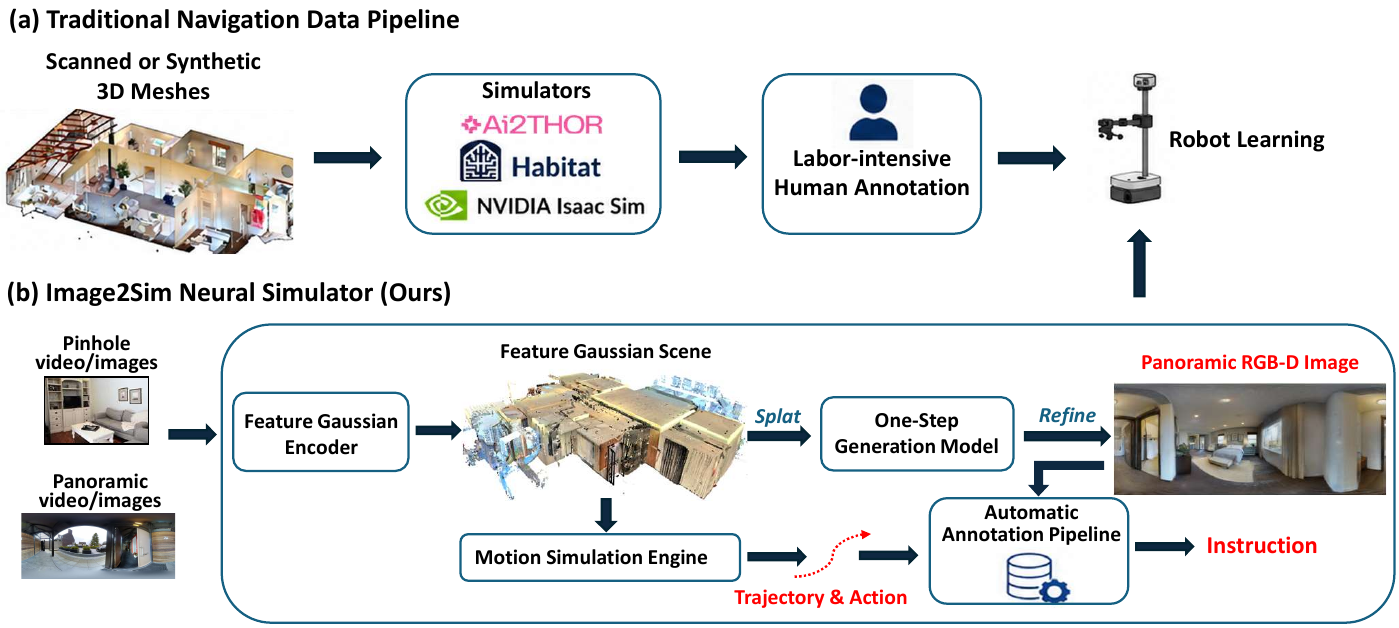}
\par\end{center}%
\end{minipage}
\caption{Comparison of (a) traditional navigation data pipeline and (b) our Image2Sim framework.}
\label{fig:Introduction}
\vspace{-10pt}
\end{figure*}

These failure modes share a common cause: existing pipelines often couple 3D geometry and visual synthesis within a single mechanism, forcing a tradeoff among scale, physical grounding, and visual fidelity. 
To overcome this tradeoff, we introduce Image2Sim, a real-time neural environment engine that constructs high-quality interactive scenes from posed RGB-D image sequences at scale. 
The key idea is to decouple 3D spatial anchoring from photorealistic observation synthesis. 
Specifically, Image2Sim first uses a feed-forward feature Gaussian model to lift observations into 3D feature Gaussians in a single pass, producing an explicit geometric and semantic scene representation. 
We further propose a Geometry-Aware One-Step Pixel Flow model to synthesize high-fidelity panoramic RGB-D observations,
which formulates rendering as a probability flow ODE conditioned on multimodal priors and guided by 3DGS alpha maps. This design enables momentum-based self-distillation and compresses multi-step generation into a direct single-step mapping. 
However, a neural renderer alone is not enough for navigation training; executable actions and aligned language instructions are further required.

Image2Sim therefore extends from a neural renderer into a fully automated embodied data engine. 
Instead of relying on unconstrained generation, we integrate a white-box, geometry-aware motion engine that respects collision constraints and produces physically valid trajectories targeting long-tail and object-centric goals. 
A large vision-language model~\cite{bai2025qwen3} then generates natural language instructions aligned with these paths. 
The framework constructs vision, action, and instruction data jointly at scale, producing over 10M navigation samples across 20K interactive scenes. 
Navigation models trained entirely within these neural environments show strong gains on standard embodied benchmarks and transfer effectively to real-world zero-shot settings. 
These results indicate that scalable neural simulation is a practical route toward embodied scaling.

In summary, our contributions are threefold:
\begin{itemize}[leftmargin=5mm]
\item \textbf{A Geometry-Aware Neural Simulator.} 
We propose a real-time simulation framework that explicitly decouples feed-forward 3D scene construction from photorealistic panoramic RGB-D rendering. 
This design enables geometrically grounded, high-fidelity, and computationally efficient environments, supporting panoramic RGB-D synthesis at approximately 40 FPS on a single RTX 4090 GPU.
\item \textbf{A Scalable Embodied Data Engine.} 
We develop a fully automated pipeline integrating object-centric target mining, physics-aware trajectory generation, and VLM-based instruction annotation. 
The pipeline produces over 10M navigation training samples across 20K interactive scenes.
\item \textbf{Empirical Evidence for Embodied Scaling.} 
We demonstrate that increasing Image2Sim-generated data yields consistent and unsaturated navigation gains, while models trained exclusively in our neural environments achieve strong cross-simulator performance and effective real-world zero-shot transfer. 
These findings validate scalable neural simulation as a practical path for improving embodied navigation.
\end{itemize}

\section{Related Work}

\paragraph{Simulation Environments for Embodied Navigation.}

Embodied navigation mainly relies on two simulation paradigms. 
Real-scanned environments such as Matterport3D, HM3D, Gibson, and Replica~\cite{chang2017matterport3d,ramakrishnan2habitat,xia2018gibson,straub2019replica}, together with simulators such as Habitat~\cite{savva2019habitat,wang2025rethinking}, provide strong geometric realism, real-world layout statistics, and physical grounding. 
However, high-quality 3D scanning and digital-twin post-processing are expensive and labor-intensive, limiting scene scale and diversity. 
Synthetic and procedural environments~\cite{deitke2022️,khanna2024habitat,kolve2017ai2,mittal2025isaac,lin2025vlnverse,ehsani2024spoc,kim2026molmospaces,yang2024holodeck,nasiriany2024robocasa} offer better scalability, controllability, and task diversity, but often suffer from sim-to-real gaps caused by unrealistic assets, layouts, and rendering statistics. 
Image2Sim instead constructs interactive environments directly from large-scale images and videos, aiming to combine scalable scene generation with the realism of real-world captures.

\paragraph{Neural Scene Representations and World Models.}

Neural simulation is closely related to neural scene representations and generative world models. 
Optimization-based methods such as NeRF~\cite{mildenhall2021nerf} and 3D Gaussian Splatting~\cite{kerbl20233d}, as well as feature-augmented variants~\cite{zhou2024feature}, achieve high-fidelity rendering but require costly per-scene optimization, limiting their use as large-scale embodied data engines. 
Recent embodied simulators built on such neural scenes~\cite{xie2025vid2sim,liu2026navgsim,miao2025towards,chhablani2025embodiedsplat,liu2025wanderland,yoo2026ready,escontrela2025gaussgym,sim2026floor} inherit this scalability limitation. 
Feed-forward reconstruction methods~\cite{charatan2024pixelsplat,chen2024mvsplat,wang2024freesplat,ye2025yonosplat,lee2025omnisplat,chen2025splatter,ziwen2025long,jiang2025anysplat} reduce scene construction cost, but often degrade under noisy inputs, partial observations, or sparse views, producing incomplete geometry and missing regions. 
Generative world models~\cite{bruce2024genie, bar2025navigation, assran2025v,chen2026imaginav} can synthesize realistic visual observations, but generally lack persistent 3D structure, explicit navigability, and collision-aware geometry for closed-loop interaction. 
Image2Sim addresses these limitations by decoupling \emph{3D spatial anchoring} from \emph{photorealistic observation synthesis}, using feed-forward feature Gaussians for persistent geometry and a one-step pixel-flow renderer for panoramic RGB-D generation.

\paragraph{Scaling Vision-Language Navigation.}

Scaling data has become a central direction in Vision-Language Navigation (VLN). 
Early progress was largely driven by human-annotated datasets~\cite{anderson2018vision,krantz2020beyond,rxr,qi2020reverie,zhu2021soon}, which provide high-quality training supervision but remain constrained by annotation cost and the limited number of underlying 3D environments. 
To reduce annotation bottlenecks, subsequent works synthesize additional instructions, goals, or trajectories using pretrained vision-language models or goal-oriented data engines~\cite{fried2018speaker,hao2020towards,kamath2023new,chen2022learning,wang2023scaling,wang2024bootstrapping,wang2025navrag,li2025learning,zhang2025embodied}. 
These methods substantially increase the amount and diversity of language-action supervision, but the resulting data is still grounded in a fixed pool of simulator-supported scenes~\cite{chang2017matterport3d,ramakrishnan2habitat,xia2018gibson}. 
As a result, the distribution of interactive environments remains comparatively narrow, leaving scene-scale diversity as an underexplored dimension of VLN scaling. 
Image2Sim addresses this dimension by converting large-scale image and video data into near 20K interactive environments, from which it generates over 10M physically grounded vision-language-action samples.

\nocite{wang2025progress,zhou2024navgpt2}

\section{Method}
\label{sec:method}

\subsection{Problem Formulation}

Let a captured scene be represented by a sequence of posed RGB-D observations
$\mathcal{O}=\{(I_i, D_i, \mathbf{K}_i, \mathbf{T}_i)\}_{i=1}^{N}$,
where $I_i \in \mathbb{R}^{H \times W \times 3}$ is an RGB image, $D_i$ is its aligned depth map, $\mathbf{K}_i$ denotes the camera intrinsics, and $\mathbf{T}_i \in SE(3)$ is the camera-to-world pose. 
For RGB-only videos, the depth, camera poses, and intrinsics can be obtained from off-the-shelf 3D foundation models~\cite{wang2025vggt,wang2025pi,lin2025depth}.

\begin{figure*}[h!]
\vspace{-8pt}
\centering
\includegraphics[width=0.95\columnwidth]{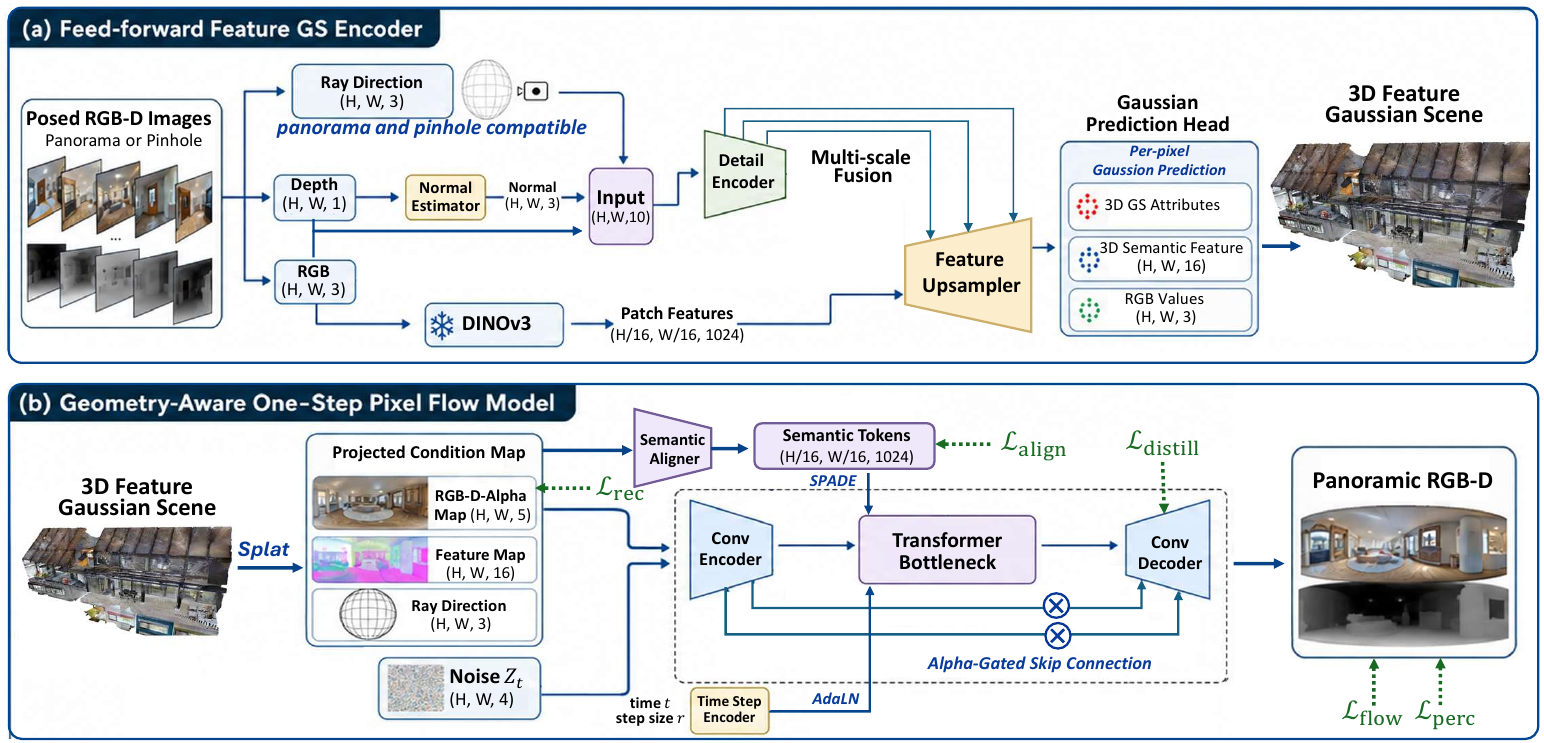}
\caption{Framework of the feed-forward GS encoder and one-step pixel flow model.}
\label{fig:Render}
\vspace{-10pt}
\end{figure*}

We build Image2Sim to convert passive scene captures into executable neural simulators for scalable embodied navigation data synthesis. 
As illustrated in Figure~\ref{fig:Render} and 
\ref{fig:Date_generation}, the system couples three components: 
(i) feed-forward construction of a persistent 3D feature-Gaussian scene, 
(ii) geometry-aware one-step panoramic RGB-D rendering, and 
(iii) executable motion simulation for trajectory and instruction generation. 
The key design principle is to combine explicit 3D grounding with generative completion: reliable observed geometry is preserved through Gaussian splatting, and unobserved regions are completed by a conditional pixel-flow renderer.

Given observations $\mathcal{O}$, Image2Sim constructs an interactive simulator $\mathcal{S}=\{\mathcal{G}, \mathcal{M}, \mathcal{R}_{\theta}\}$, where $\mathcal{G}$ is a persistent 3D feature-Gaussian scene, $\mathcal{M}$ is a traversable voxel connectivity graph, and $\mathcal{R}_{\theta}$ maps arbitrary agent poses to panoramic RGB-D observations. 
$\mathcal{S}$ supports physically grounded atomic actions, collision-aware trajectory sampling, and panoramic rendering along feasible paths. Using $\mathcal{S}$, we synthesize aligned vision-language-action triplets $\mathcal{D}_{\mathrm{nav}} = \{(X_{1:T}, a_{1:T}, Y)\}$, where $X_{1:T}$ are rendered panoramic observations, $a_{1:T}$ are executable low-level actions, and $Y$ denotes the corresponding navigation instruction.

\subsection{Feed-Forward 3D Scene Construction}

\noindent \textbf{Semantic-geometric pixel lifting.}
Image2Sim first lifts posed RGB-D frames into a persistent 3D representation without per-scene optimization. 
For each frame, we extract dense visual features using a dual-stream encoder. 
A frozen DINOv3~\cite{simeoni2025dinov3} backbone provides high-level semantic features, and a lightweight geometric detail stream preserves local details from RGB, metric depth, surface normals, and ray directions. 
The ray-direction encoding makes the Gaussian encoder compatible with both pinhole and panoramic cameras.

The semantic and geometric features are fused by a feature upsampler and decoded by a Gaussian prediction head into dense 3D feature Gaussians:
\begin{equation}
    \mathcal{G}=\{g_j\}_{j=1}^{M}, 
    \qquad
    g_j=(\boldsymbol{\mu}_j,\mathbf{s}_j,\mathbf{q}_j,\alpha_j,\mathbf{c}_j,\mathbf{f}_j),
\end{equation}
where $\boldsymbol{\mu}_j \in \mathbb{R}^{3}$ is the Gaussian center, $\mathbf{s}_j \in \mathbb{R}^{3}$ is the anisotropic scale, $\mathbf{q}_j \in \mathbb{R}^{4}$ is the rotation, $\alpha_j \in [0,1]$ is the opacity, $\mathbf{c}_j \in \mathbb{R}^{3}$ stores RGB values, and $\mathbf{f}_j$ stores compact semantic features. 
The Gaussian centers are deterministically unprojected from depth and voxel-downsampled, which anchors the representation to metric scene geometry.
Unlike optimization-based Gaussian reconstruction, this feed-forward lifting amortizes scene construction across environments and enables real-time conversion of large image collections into executable 3D scenes.

\noindent \textbf{Projection and semantic grounding.}
Given a target panoramic pose $p$, the Gaussian scene $\mathcal{G}$ can be splatted into the panorama coordinate system to obtain projected RGB, depth, feature, and opacity maps:
\begin{equation}
    \mathbf{C}_{p}
    =
    [\tilde{\mathbf{I}}_{p}, 
     \tilde{\mathbf{D}}_{p}, 
     \tilde{\mathbf{F}}_{p}, 
     \tilde{\mathbf{A}}_{p}, 
     \mathbf{R}_{p}],
\end{equation}
where $\tilde{\mathbf{I}}_{p}$, $\tilde{\mathbf{D}}_{p}$, and $\tilde{\mathbf{F}}_{p}$ are the splatted RGB, depth, and semantic feature maps, $\tilde{\mathbf{A}}_{p}$ is the accumulated opacity map, and $\mathbf{R}_{p}$ is the panoramic ray-direction map. 

Observed regions are supervised by an RGB-D reconstruction loss. Projected feature maps are then tokenized by a semantic aligner $\Phi$ and aligned with DINOv3 features extracted from the target RGB panorama:
\vspace{-10pt}
\begin{equation}
    \mathcal{L}_{\mathrm{rec}}
    =
    \left(
    \| \tilde{\mathbf{I}}_p - \mathbf{I}_p \|_1
    +
    \| \tilde{\mathbf{D}}_p - \mathbf{D}_p \|_1
    \right)
    \odot \mathbf{M}_p, \quad
    \mathcal{L}_{\mathrm{align}}
    =
    \left(
    1 -
    \frac{
    \Phi(\mathbf{C}_p) \cdot \mathbf{S}_p
    }{
    \| \Phi(\mathbf{C}_p) \|_2
    \| \mathbf{S}_p \|_2
    }
    \right)
    \odot \mathbf{M}'_p .
\end{equation}
$\mathbf{M}_p$ is a validity mask derived from ground-truth depth, $\mathbf{S}_p$ denotes ground-truth DINOv3 tokens from the target RGB image, and $\mathbf{M}'_p$ is the downsampled validity mask. This supervision encourages $\mathcal{G}$ to encode local appearance and semantic layout cues that are useful for novel-view completion.

\subsection{Geometry-Aware One-Step Pixel Flow Rendering}

Direct Gaussian splatting preserves observed geometry but produces holes and noisy projections under sparse captures. 
Conversely, unconstrained image generation can synthesize plausible pixels but lacks persistent 3D grounding. 
Image2Sim bridges this gap with a geometry-aware pixel-flow renderer that performs conditional generative completion under explicit 3D constraints.

\noindent \textbf{Alpha-gated source state.}
Let $\mathbf{X}_p=[\mathbf{I}_p,\mathbf{D}_p]$ denote the target panoramic RGB-D observation at pose $p$. 
We first obtain a projected RGB-D prior: $\tilde{\mathbf{X}}_p = [\tilde{\mathbf{I}}_p,\tilde{\mathbf{D}}_p]$ and an opacity map $\tilde{\mathbf{A}}_p$ that estimates the reliability of the 3D projection. We then define a spatially varying noise scale:
\begin{equation}
    \Sigma(\tilde{\mathbf{A}}_p)
    =
    \tilde{\mathbf{A}}_p \cdot \sigma_{\mathrm{small}}
    +
    (1-\tilde{\mathbf{A}}_p) \cdot \sigma_{\mathrm{large}}.
\end{equation}
Finally, the alpha-gated source state is constructed as:
\begin{equation}
    \mathbf{z}_{\mathrm{src}}
    =
    \boldsymbol{\tilde{\mathbf{A}}}_p \odot \tilde{\mathbf{X}}_p
    +
    \Sigma(\boldsymbol{\tilde{\mathbf{A}}}_p)\odot\boldsymbol{\epsilon},
    \qquad
    \boldsymbol{\epsilon}\sim\mathcal{N}(0,\mathbf{I}).
\end{equation}
High-opacity regions remain close to the projected 3D evidence. In contrast, low-opacity regions receive larger perturbations and are handled by generative completion.

\noindent \textbf{Single-step pixel-flow renderer.}
To estimate the geometry-conditioned probability flow, we parameterize the denoised prediction $\mathbf{x}_{\theta}$ with a UNet-style architecture. 
A convolutional encoder first compresses the alpha-gated source state $\mathbf{z}_{\mathrm{src}}$ together with the geometric condition $\mathbf{C}_{p}$. 
A deep transformer bottleneck then captures global layout dependencies and infers unobserved scene structures. 
Within this bottleneck, time step is injected through AdaLN~\cite{peebles2023scalable}, and semantic tokens $\Phi(\mathbf{C}_p)$ are injected through SPADE-like spatially-adaptive normalization~\cite{park2019semantic} to provide high-level layout guidance. 
During convolutional upsampling, multi-scale skip connections are further modulated by an alpha pyramid derived from $\tilde{\mathbf{A}}_{p}$. 
These alpha-gated skips preserve reliable projected evidence in high-opacity regions and suppress noisy features in missing regions.

\subsection{Training the Pixel-Flow Renderer}

To generate high-quality panoramic RGB-D images in real time, we replace traditional multi-step flow matching~\cite{lipman2022flow} with a continuous-time MeanFlow formulation~\cite{geng2025mean,lu2026one}, which trains the network to estimate the average velocity for one-step transport from the source state to the target observation.

During training, we hold out panoramic RGB-D frames $\mathbf{X}_{p}$ as target supervision. 
Given the source state $\mathbf{z}_{\mathrm{src}}$, we define a linear probability path following the pixel MeanFlow convention~\cite{lu2026one}, where $t=0$ corresponds to the target panorama and $t=1$ corresponds to the projected noisy source:
\begin{equation}
    \mathbf{z}_t
    =
    (1-t)\mathbf{X}_{p}
    +
    t\mathbf{z}_{\mathrm{src}},
    \qquad
    \mathbf{v}_{\mathrm{target}}
    =
    \mathbf{z}_{\mathrm{src}}
    -
    \mathbf{X}_{p},
    \qquad
    t\in[0,1].
\end{equation}

Following pixel MeanFlow~\cite{lu2026one}, we let the network predict the denoised observation $\mathbf{x}_{\theta}$ and parameterize the average velocity as $\mathbf{u}_{\theta} = (\mathbf{z}_t - \mathbf{x}_{\theta}) / t$. We then construct the instantaneous velocity field $\mathbf{V}_{\theta}$ and flow loss:

\vspace{-10pt}
\begin{equation}
    \mathbf{u}_{\theta} = \frac{\mathbf{z}_t - \mathbf{x}_{\theta}}{\max(t,\varepsilon)},\ 
    \mathbf{V}_{\theta}
    =
    \mathbf{u}_{\theta}
    +
    (t-r)
    \frac{d\mathbf{u}_{\theta}}{dt}, \ \
    \mathcal{L}_{\mathrm{flow}}
    =
    \mathbb{E}_{r,t,\boldsymbol{\epsilon},\mathbf{X}_{p}}
    \left[
    \mathbf{W}(\tilde{\mathbf{A}}_{p})\odot
    \left\|
    \left(
    \mathbf{V}_{\theta}
    -
    \mathbf{v}_{\mathrm{target}}
    \right)
    \right\|_2^2
    \right].
\end{equation}
Here, $r$ and $t$ define the integration interval, $\varepsilon$ is a small constant to prevent division by zero. 
The total derivative $\frac{d\mathbf{u}_{\theta}}{dt}$ captures how the predicted average velocity changes along the probability path. 
Instead of computing this derivative with expensive higher-order backpropagation, we use forward-mode Jacobian-vector products (JVPs)~\cite{geng2025mean} to efficiently evaluate the directional derivative along the flow trajectory, and apply stop-gradient to this derivative term in the loss. 
The opacity-adaptive weighting map $\mathbf{W}(\tilde{\mathbf{A}}_{p})$ assigns larger gradients to low-confidence regions where $\tilde{\mathbf{A}}_{p}\approx 0$. 
This explicitly concentrates the learning signal on unobserved regions that require generative completion while concurrently preventing the network from unnecessarily modifying reliable 3D projections.

\noindent \textbf{Momentum-based self-distillation.}
Single-step generation is efficient, but can be unstable in severely occluded or weakly observed regions. 
Inspired by DINO~\cite{caron2021emerging}, we introduce a momentum-based self-distillation loss to improve stability and suppress hallucinated artifacts. We maintain a teacher network updated by the exponential moving average (EMA) of the student parameters. 
Unlike the student that receives the noisy alpha-gated source state, the teacher is conditioned on a privileged source state $\mathbf{z}_{\mathrm{gt}}$ constructed from the ground-truth RGB-D panorama. 
The student is then encouraged to match the 
multi-scale decoder features of the teacher across a set of decoder levels $\mathcal{K}$:
\begin{equation}
    \mathcal{L}_{\mathrm{distill}}
    =
    \sum_{k \in \mathcal{K}}
    \tilde{w}_k
    \left[
    1 -
    \frac{
    \mathbf{f}_k^{S} \cdot \mathbf{f}_k^{T}
    }{
    \|\mathbf{f}_k^{S}\|_2
    \|\mathbf{f}_k^{T}\|_2
    }
    +
    \beta
    \|\mathbf{f}_k^{S}-\mathbf{f}_k^{T}\|_2^2
    \right],
\end{equation}
where $\mathbf{f}_k^{S}$ and $\mathbf{f}_k^{T}$ are the student and teacher feature maps at level $k$, respectively, and $\tilde{w}_k$ are normalized layer-wise weights that emphasize deeper semantic layers. 
This distillation improves single-step completion ability by transferring the cleaner latent structure of the teacher to the student.

\noindent \textbf{Full learning objective.}
The complete training objective jointly optimizes geometric projection, semantic grounding, generative completion, and single-step rendering stability:
$\mathcal{L}_{\mathrm{total}}
=
\lambda_{\mathrm{rec}}\mathcal{L}_{\mathrm{rec}}
+
\lambda_{\mathrm{align}}\mathcal{L}_{\mathrm{align}}
+
\lambda_{\mathrm{flow}}\mathcal{L}_{\mathrm{flow}}
+
\lambda_{\mathrm{distill}}\mathcal{L}_{\mathrm{distill}}
+
\lambda_{\mathrm{perc}}\mathcal{L}_{\mathrm{perc}}$,
where $\mathcal{L}_{\mathrm{perc}}$ is a perceptual LPIPS loss~\cite{zhang2018unreasonable}, and the $\lambda$ coefficients balance the different terms.

\subsection{Motion Simulation and Trajectory Generation}

\begin{figure*}[h!]
\noindent\begin{minipage}[H]{1\columnwidth}%
\begin{center}
\includegraphics[width=0.9\columnwidth]{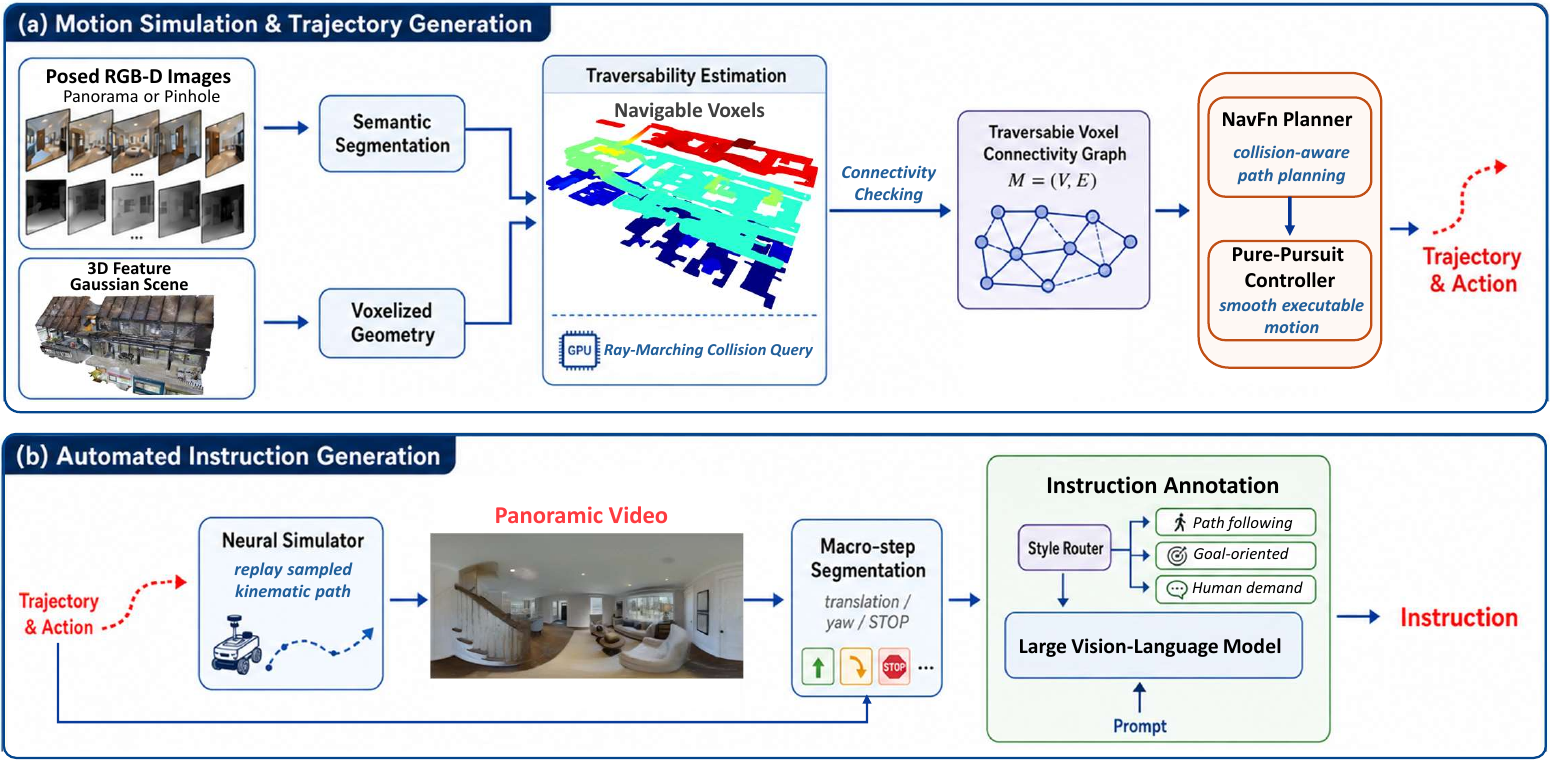}
\par\end{center}%
\end{minipage}
\caption{Framework of the motion simulation engine and automated instruction generation.}
\label{fig:Date_generation}
\vspace{-10pt}
\end{figure*}

High-fidelity visual rendering alone is insufficient for embodied learning: the simulator must also support physically valid agent motion. 
As shown in Figure~\ref{fig:Date_generation}, Image2Sim addresses this requirement by deriving a dense navigability structure directly from the reconstructed 3D scene. 
We voxelize the Gaussian scene into a dense grid $\mathcal{V}$ and determine traversability using both semantic labels and geometric obstacle constraints, such as floor height, free-space connectivity, and clearance from occupied regions. 
To account for finite-size agents, we perform GPU-parallel ray marching over the voxelized geometry to query occupancy within the agent footprint. 
This allows the simulator to reject invalid motions, detect geometric penetration, and support physically plausible sliding along obstacle boundaries. 
The resulting structure forms a traversable voxel connectivity graph
$\mathcal{M}=(\mathcal{V},\mathcal{E})$
where nodes denote traversable voxels and edges connect locally reachable neighboring states. 

For trajectory generation, we plan paths over $\mathcal{M}$ using path-length and collision-aware edge costs, similar to the NavFn planner~\cite{macenski2020marathon,macenski2023desks}. 
The collision-aware term penalizes states close to obstacles, encouraging the planner to favor safer corridors rather than merely shortest paths. 
A lookahead pure-pursuit controller~\cite{macenski2023regulated} then converts discrete graph paths into smooth executable motion. 
Finally, the simulator replays these trajectories to render temporally consistent panoramic RGB-D observations along the exact kinematic path.

\subsection{Automated Instruction Generation}
 
Image2Sim further converts simulated trajectories into vision-language-action annotations. 
Each sampled feasible trajectory is replayed in the neural simulator to render panoramic RGB-D observations along the executed motion. 
To bridge the granularity gap between dense low-level actions and human navigation language, we segment each trajectory into semantic macro-steps according to accumulated translation, yaw rotation, and terminal STOP events. 
Each macro-step is mapped to a motion primitive and aligned with the panorama rendered at its starting pose, yielding a compact observation-motion sequence for language annotation. 
The aligned sequence is then annotated by Qwen3-VL-32B-Instruct~\cite{bai2025qwen3} to synthesize natural-language navigation instructions. 
A style-conditioned annotation router further diversifies the instructions across fine-grained path following, high-level goal-oriented descriptions, and natural human demands. 
This automatic pipeline produces scalable embodied navigation data without manual trajectory or instruction annotation.

\section{Experiments}
\label{sec:experiments}

\subsection{Experimental Setup}

\noindent \textbf{Datasets.}
We build interactive environments from 19,936 real-world and synthetic scenes (training only, excluding evaluation/test scenes) with images or videos, including RealSee3D~\cite{Li2025realsee3d_data}, Structured3D~\cite{Structured3D}, ARKitScenes~\cite{baruch1arkitscenes}, HM3D~\cite{ramakrishnan2habitat}, ScanNet~\cite{dai2017scannet}, Gibson~\cite{xia2018gibson}, and Matterport3D~\cite{chang2017matterport3d}. Panoramic inputs are fixed at $512{\times}1024$, while pinhole images are resized to suitable multiples of 16. The pixel-flow renderer outputs $512{\times}1024$ panoramic RGB-D images. From these 20K interactive environments, our data engine synthesizes over 10M vision-language-action samples for navigation training. More dataset details are provided in the Appendix.

\noindent \textbf{Image2Sim Neural Simulator.}
Image2Sim is trained on Matterport3D-360~\cite{chang2017matterport3d,rey2022360monodepth}, RealSee3D~\cite{Li2025realsee3d_data}, and Structured3D~\cite{Structured3D} for 400K iterations using 8 A6000 Ada GPUs with a per-GPU batch size of 2. We adopt a three-stage curriculum: geometry initialization of Gaussian encoder in the first 100K steps, pixel-flow training with LPIPS~\cite{zhang2018unreasonable} and momentum-based self-distillation after 100K steps, and high-frequency refinement after 300K steps by freezing the convolutional encoder and transformer bottleneck. The self-distillation teacher is updated with EMA~\cite{caron2021emerging} decay 0.999.

\noindent \textbf{Image2Nav Navigation Model.}
Image2Nav is our baseline navigation model, built on Qwen3-VL-4B-Instruct~\cite{bai2025qwen3} and predicts two discrete actions per step from interleaved historical RGB observations and executed actions. We train and evaluate Image2Nav under two monocular observation settings: the standard pinhole camera ($336{\times}336$, $90^\circ$ HFOV/VFOV) and the ultra-wide equirectangular image ($512{\times}336$, $180^\circ$ HFOV and $118^\circ$ VFOV). Unless otherwise specified, we use the standard pinhole camera setting for experimental analyses. Within the training input, each trajectory contains up to 32 uniformly sampled RGB observations, paired with 64 action labels. For long-horizon inference, we use a streaming sparse-dense context: 24 uniformly sampled historical frames provide global memory, and an 8-frame recent window supports dense local prediction with KV-cache reuse. Image2Nav is trained on 64 H20 GPUs with DeepSpeed ZeRO-2~\cite{rasley2020deepspeed}, bfloat16 precision, and gradient checkpointing, fine-tuning only the LLM and MLP projector. Stage 1 performs 70K-step offline behavior cloning, and Stage 2 performs 10K-step online DAgger~\cite{ross2011reduction,chen2022think} in Image2Sim with randomized 4--32 step chunks that mix model-predicted and oracle actions while supervising all executed spans with oracle labels.

\subsection{Novel-View Rendering Evaluation}

We first evaluate the novel-view rendering quality and speed across varying pose and depth noise distributions (Table~\ref{tab:combined_results}). 
The pure generative baselines, DiT360~\cite{feng2025dit360} and SE3DS~\cite{koh2023simple}, can produce plausible pixels but do not satisfy the real-time requirement for closed-loop interaction (0.3 FPS and 3.4 FPS). 
Conversely, feed-forward 3DGS predictors such as AnySplat~\cite{jiang2025anysplat} operate at very high frequencies ($>$115 FPS), but suffer substantial structural degradation when large regions are unobserved, as reflected by clear drops in PSNR~\cite{wang2004image} and SSIM.

\begin{table*}[htbp]
\centering
\caption{Novel-view rendering quality on datasets with different noise levels.}
\label{tab:combined_results}
\vspace{-5pt}
\resizebox{\textwidth}{!}{
\begin{tabular}{l cccc cccc cccc}
\toprule
\multirow{2}{*}{Method} & \multicolumn{4}{c}{RealSee3D-Synthesis~\cite{Li2025realsee3d_data} (No Noise)} & \multicolumn{4}{c}{Matterport3D-360~\cite{chang2017matterport3d,rey2022360monodepth} (Middle Noise)} & \multicolumn{4}{c}{RealSee3D-Real~\cite{Li2025realsee3d_data} (High Noise)} \\
\cmidrule(lr){2-5} \cmidrule(lr){6-9} \cmidrule(lr){10-13}
& LPIPS$\downarrow$ & PSNR$\uparrow$ & SSIM$\uparrow$ & \textbf{FPS$\uparrow$} 
& LPIPS$\downarrow$ & PSNR$\uparrow$ & SSIM$\uparrow$ & \textbf{FPS$\uparrow$} 
& LPIPS$\downarrow$ & PSNR$\uparrow$ & SSIM$\uparrow$ & \textbf{FPS$\uparrow$} \\
\midrule

DiT360~\cite{feng2025dit360} (Pano input)
& \textcolor{gray}{0.259} & \textcolor{gray}{24.56} & \textcolor{gray}{0.724} & \textcolor{gray}{0.3} 
& \textcolor{gray}{0.337} & \textcolor{gray}{21.63} & \textcolor{gray}{0.545} & \textcolor{gray}{0.3} 
& \textcolor{gray}{0.364} & \textcolor{gray}{19.58} & \textcolor{gray}{0.564} & \textcolor{gray}{0.3} \\
\midrule

AnySplat~\cite{jiang2025anysplat} (Pinhole input) 
& 0.378 & 20.31 & 0.555 & \textbf{125.8} 
& 0.450 & 18.32 & 0.508 & \textbf{115.1} 
& 0.582 & 15.23 & 0.415 & \textbf{137.3} \\

SE3DS~\cite{koh2023simple} (Pano input)
& 0.379 & 18.23 & 0.538 & 3.4 
& \textbf{0.426} & 18.76 & \textbf{0.536} & 3.3 
& \textbf{0.421} & 16.78 & 0.457 & 3.4 \\

\textbf{Image2Sim (Pinhole input)} 
& \underline{0.345} & \underline{23.56} & \underline{0.583} & 41.1 
& 0.432 & \textbf{20.21} & \underline{0.525} & 37.1 
& 0.488 & \underline{17.17} & \underline{0.469} & 43.0 \\

\textbf{Image2Sim (Pano input)}
& \textbf{0.335} & \textbf{23.88} & \textbf{0.591} & \underline{41.3} 
& \underline{0.428} & \underline{20.15} & 0.521 & \underline{37.5} 
& \underline{0.473} & \textbf{17.43} & \textbf{0.470} & \underline{45.6} \\

\bottomrule
\end{tabular}
}
\end{table*}

\begin{figure*}[h!]
\vspace{-15pt}
\noindent\begin{minipage}[H]{1\columnwidth}%
\begin{center}
\includegraphics[width=0.9\columnwidth]{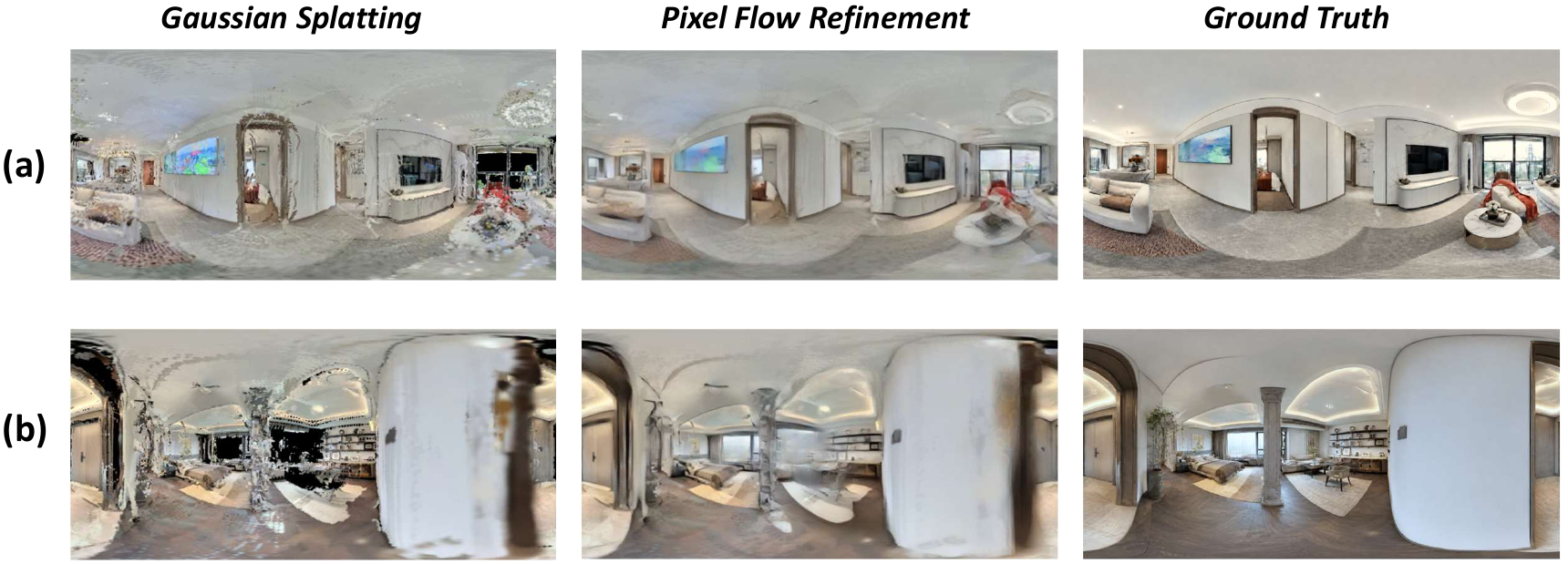}
\par\end{center}%
\end{minipage}
\caption{Visualization of Gaussian Splatting and Pixel Flow renderings in high-noise sparse scenes.}
\label{fig:visualization}
\vspace{-15pt}
\end{figure*}

By explicitly decoupling spatial anchoring from generative completion, Image2Sim achieves a more balanced tradeoff between fidelity, robustness, and efficiency. 
On RealSee3D-Real~\cite{Li2025realsee3d_data}, which contains sparse and high-noise LiDAR depth, our panoramic variant maintains 17.43 PSNR and 0.470 SSIM while running at 45.6 FPS, substantially outperforming Gaussian-only AnySplat in rendering quality. 
These results show that the geometry-aware pixel-flow model can complete structural blind spots under noisy reconstruction, as shown in Figure~\ref{fig:visualization}, while the Gaussian projection and alpha guidance help prevent unconstrained hallucination. 
The consistent performance of both pinhole and panoramic inputs further indicates that Image2Sim can accommodate heterogeneous capture formats, which is important for scaling neural simulation from diverse real-world videos and images.

\subsection{Cross-Simulator Zero-Shot Navigation Generalization}

To assess the quality of the navigation data generated by Image2Sim, we evaluate Image2Nav on the R2R-CE~\cite{krantz2020beyond}, RxR-CE~\cite{rxr}, and REVERIE-CE~\cite{qi2020reverie,wangdynam3d} benchmarks (Table~\ref{tab:comprehensive_vln_results}). 
All compared baselines are trained and evaluated within the same Habitat simulator~\cite{savva2019habitat}. 
In contrast, Image2Nav is trained \emph{exclusively} in our Image2Sim framework and evaluated \textit{zero-shot} in the Habitat simulator. 
This setting directly tests whether Image2Sim-generated data transfers across simulator domains rather than only improving in-domain performance.

\begin{table*}[htbp]
\caption{Navigation performance with monocular observations on R2R-CE, RxR-CE, and REVERIE-CE. For our Image2Nav, we compare the standard pinhole camera (336$\times$336, 90° HFOV/VFOV) and the ultra-wide equirectangular image (512$\times$336, 180° HFOV$\times$118° VFOV) settings.}
\label{tab:comprehensive_vln_results}
\vspace{-5pt}
\centering
\resizebox{\textwidth}{!}{
\begin{tabular}{lcc cccc cccc cccc}
\toprule
\multirow{2}{*}{\textbf{Methods}} & \multirow{2}{*}{\textbf{RGB}} & \multirow{2}{*}{\textbf{Depth}} & \multicolumn{4}{c}{\textbf{R2R-CE Val Unseen}} & \multicolumn{4}{c}{\textbf{RxR-CE Val Unseen}} & \multicolumn{4}{c}{\textbf{REVERIE-CE Val Unseen}} \\
\cmidrule(lr){4-7} \cmidrule(lr){8-11} \cmidrule(lr){12-15}
& & & NE$\downarrow$ & OSR$\uparrow$ & SR$\uparrow$ & SPL$\uparrow$ & NE$\downarrow$ & SR$\uparrow$ & SPL$\uparrow$ & nDTW$\uparrow$ & NE$\downarrow$ & OSR$\uparrow$ & SR$\uparrow$ & SPL$\uparrow$ \\
\midrule
NaVid~\cite{zhang2024navid}  & \checkmark & & 5.47 & 49.1 & 37.4 & 35.9 & - & - & - & - & 6.74 & 36.3 & 26.6 & 20.8 \\
Uni-NaVid~\cite{zhang2024uninavid}      & \checkmark & & 5.58 & 53.3 & 47.0 & 42.7 & 6.24 & 48.7 & 40.9 & - & - & - & - & - \\
NaVILA~\cite{cheng2024navila}  & \checkmark & & 5.22 & 62.5 & 54.0 & 49.0 & 6.77 & 49.3 & 44.0 & 58.8 & - & - & - & - \\
Dynam3D~\cite{wangdynam3d}  & \checkmark & \checkmark & 5.34 & 62.1 & 52.9 & 45.7 & - & - & - & - & 6.22 & 48.9 & 40.1 & 28.5 \\
NavFoM~\cite{zhang2025embodied}    & \checkmark &  & 5.01 & 64.9 & 56.2 & 51.2 & 5.51 & 57.4 & 49.4 & 60.2 & - & - & - & - \\
StreamVLN~\cite{wei2025streamvln}      & \checkmark & \checkmark & 4.98 & 64.2 & 56.9 & 51.9 & 6.22 & 52.9 & 46.0 & 61.9 & - & - & - & - \\
D3D-VLP~\cite{wang2025d3d}    & \checkmark & \checkmark & 4.73 & 67.2 & 61.3 & 56.1 & - & - & - & - &  5.36 & 56.9 & 47.5 & 34.7 \\
JanusVLN~\cite{zeng2025janusvln}    & \checkmark & & 4.78 & 65.2 & 60.5 & 56.8 & 6.06 & 56.2 & 47.5 & 62.1 & - & - & - & - \\
InternVLA-N1~\cite{wang2025internvla} & \checkmark & \checkmark & 4.83 & 63.3 & 58.2 & 54.0 & 5.91 & 53.5 & 46.1 & 65.3 & - & - & - & - \\
EfficientVLN~\cite{zheng2025efficient}   & \checkmark & & 4.18 & 73.7 & 64.2 & 55.9 & 3.88 & 67.0 & 54.3 & 68.4 & - & - & - & - \\
DualVLN~\cite{wei2025ground} & \checkmark & & 4.05 & 70.7 & 64.3 & 58.5 & 4.58 & 61.4 & 51.8 & 70.0 & - & - & - & - \\
\midrule
\textbf{Image2Nav (90° FOV)}& \checkmark & & 3.96 & 72.9 & 66.3 & 61.5 & 4.03 & 65.1 & 54.7 & 70.5 & \textbf{4.90} & 59.2 & 49.1 & 37.0 \\

\textbf{Image2Nav (180° FOV)}& \checkmark & & \textbf{3.71} & \textbf{76.1} & \textbf{70.3} & \textbf{65.6} & \textbf{3.74} & \textbf{70.7} & \textbf{59.1} & \textbf{71.8} & 5.08 & \textbf{59.5} & \textbf{53.7} & \textbf{42.7} \\
\bottomrule
\end{tabular}
}
\end{table*}

Despite the substantial domain shift, Image2Nav achieves new state-of-the-art results and surpasses strong in-domain methods such as EfficientVLN and DualVLN. 
The improvements are consistent across path-following benchmarks and object-centric REVERIE-CE, suggesting that the generated data provides useful supervision for both instruction following and goal-oriented navigation. 
This result indicates that Image2Sim does not merely synthesize visually plausible observations but produces physically executable trajectories and language-action supervision that help the policy learn generalizable navigation behaviors rather than overfitting to simulator-specific biases.

\subsection{Scaling Law for Navigation Training Data}

\begin{wrapfigure}{l}{0.5\textwidth}
    \centering
    \vspace{-5pt}
    \captionof{table}{\small{R2R-CE metrics on different dataset scales.}}
    \label{tab:image2nav_scaling}
    \vspace{-5pt}
    \resizebox{\linewidth}{!}{
        \begin{tabular}{l cccc}
            \toprule
            Training Data & NE & OSR & SR & SPL \\
            \midrule
            R2R, RxR & 5.53 & 53.8 & 46.1 & 41.3 \\
            R2R, RxR, Image2Sim-\textbf{1M} & 4.87 & 65.4 & 56.9 & 49.5 \\
            R2R, RxR, Image2Sim-\textbf{5M} & 4.18 & 70.0 & 63.8 & 55.9 \\
            R2R, RxR, Image2Sim-\textbf{10M} & \textbf{3.96} & \textbf{72.9} & \textbf{66.3} & \textbf{61.5} \\
            \bottomrule
        \end{tabular}
    }

    \vspace{-2pt} 
    
    \includegraphics[width=\linewidth]{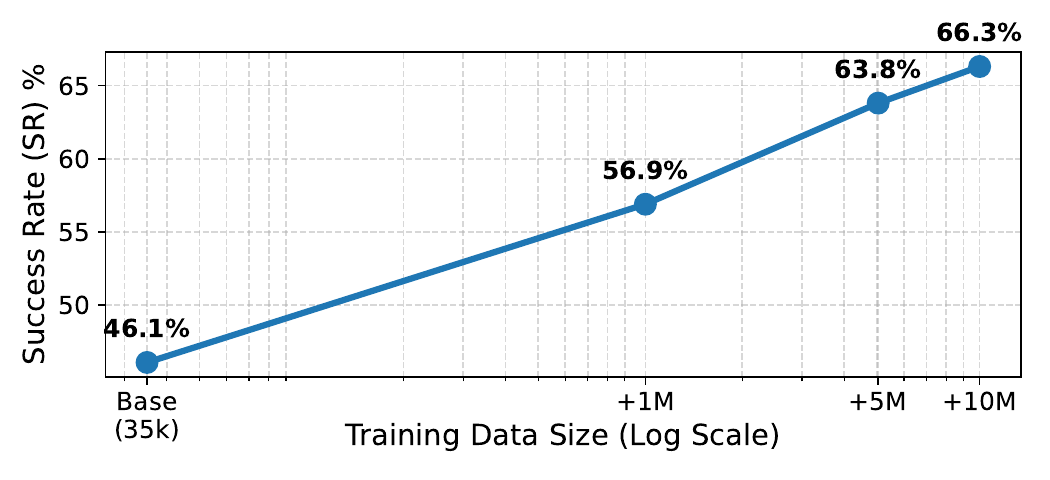}
    \vspace{-15pt}
    \captionof{figure}{Scaling law curve of training data.}
    \vspace{-20pt}\label{fig:scaling_law}
\vspace{-10pt}
\end{wrapfigure}

A central goal of Image2Sim is to alleviate the environmental data bottleneck in embodied navigation. 
As shown in Table~\ref{tab:image2nav_scaling} and Figure~\ref{fig:scaling_law}, we transfer the human-annotated R2R and RxR data from habitat into Image2Sim, then scale up with generated samples.
Scaling training samples to 10M brings consistent and unsaturated gains across all metrics, revealing a clear log-linear trend: SR improves from 46.1 to 66.3, while SPL increases from 41.3 to 61.5. 
These results suggest that current navigation generalization is still strongly data-limited. 
Importantly, the gains are not merely due to instruction augmentation over a fixed environment set. 
As the training scale increases, Image2Sim jointly expands the number of interactive scenes, feasible trajectories, rendered observations, and generated instructions, exposing the policy to broader layout, visual, action, and language distributions. 
This coupled scaling of environments and annotations supports Image2Sim as a scalable embodied data engine for large-scale navigation training.

\vspace{-10pt}
\subsection{Ablation Study}

We analyze the physical and geometric principles underlying our pixel-flow architecture in Table~\ref{tab:ablation}. 
Using only the 3DGS projection (w/o pixel flow) severely degrades rendering quality, confirming that deterministic splatting cannot overcome the inherent sparsity and holes of feed-forward reconstructions. 
The removal of the projected semantic feature map causes clear performance drops, indicating that high-level semantics are necessary to constrain generative completion in unobserved regions. 
Without alpha-gated fusion, the model no longer explicitly separates reliable projected evidence from uncertain missing regions, which can overwrite valid 3D geometry and reduce reconstruction fidelity. 
Finally, removing momentum-based self-distillation also degrades performance, showing that the EMA teacher helps stabilize one-step generation and suppress artifacts in weakly observed regions.

\begin{table}[ht!]
\vspace{-10pt}
  \centering 
  
  \begin{minipage}{0.48\textwidth}
    \flushleft 
    \caption{Ablation study of pixel flow model.}
    \label{tab:ablation}
    \setlength{\tabcolsep}{2pt} 
    \footnotesize
    \begin{tabular}{lcccc}
      \toprule
      Settings & LPIPS $\downarrow$ & PSNR $\uparrow$ & SSIM $\uparrow$ & FPS $\uparrow$ \\
      \midrule
      w/o pixel flow model   & 0.518 & 17.75 & 0.438 & \textbf{80.3} \\
      w/o semantic features  & 0.449 & 18.36 & 0.459 & 42.9 \\
      w/o alpha-gated fusion & 0.447 & 18.57 & 0.506 & 37.5 \\
      w/o self-distillation  & 0.453 & 19.84 & 0.518 & 37.5 \\
      \midrule 
      All settings           & \textbf{0.428} & \textbf{20.15} & \textbf{0.521} & 37.5 \\
      \bottomrule
    \end{tabular}
  \end{minipage}
  \hfill 
  \begin{minipage}{0.5\textwidth}
    \centering
    \caption{Real-world Navigation Performance.}
    \label{tab:navigation_results}
    \setlength{\tabcolsep}{4pt} 
    \footnotesize
    \begin{tabular}{lcccc}
      \toprule
      \multirow{2}{*}{Methods} & \multicolumn{2}{c}{Path-follow} & \multicolumn{2}{c}{Goal-orient} \\
      \cmidrule(lr){2-3} \cmidrule(lr){4-5}
      & OSR & SR & OSR & SR \\
      \midrule
      JanusVLN  & 12/20 & 8/20 & 5/20 & 4/20 \\
      DualVLN   & 13/20 & 8/20 & 8/20 & 5/20 \\
      Image2Nav & \textbf{14/20} & \textbf{11/20} & \textbf{13/20} & \textbf{9/20} \\
      \bottomrule
    \end{tabular}
  \end{minipage}
\vspace{-10pt}
\end{table}

\subsection{Real-World Navigation Evaluation}

Real-world experiments are conducted on the Hello Robot Stretch 3 in a home-like environment, with inference deployed on a workstation equipped with a RTX 4090 GPU. As shown in Table~\ref{tab:navigation_results}, 
each method is tested on 20 trials under path-following instruction and goal-oriented instruction. 
Image2Nav achieves the best performance in both scenarios, improving path-following SR from 8/20 to 11/20 and goal-oriented SR from 5/20 to 9/20 over the strongest baselines. These results suggest that Image2Sim-generated data improves transfer beyond simulator benchmarks, especially for goal-oriented navigation. The consistent improvement supports our main claim that scalable, physically executable, and visually diverse neural simulation can help reduce the sim-to-real gap for embodied navigation.

\section{Conclusion}

We introduce Image2Sim, a generative neural simulator that converts passive visual captures into scalable, physically grounded, and interactive environments for embodied navigation. By combining feed-forward 3D feature-Gaussian reconstruction, geometry-aware one-step panoramic RGB-D generation, collision-aware motion simulation, and automated instruction generation, Image2Sim enables large-scale synthesis of vision-language-action data without manual scene modeling or trajectory annotation. Experiments show that navigation models trained with Image2Sim achieve strong benchmark performance and improved cross-simulator and real-world generalization, suggesting neural simulation as a practical foundation for scaling embodied navigation.
\vspace{-10pt}

\paragraph{Limitations.}
Several limitations still remain. 
First, to support real-time interaction, low computational cost, and online DAgger/RL training, our compact renderer trades off model capacity and generation ability. 
Second, the simulator mainly handles navigation-level physical validity; richer contact dynamics, movable objects, and human--robot interactions are not yet supported. 
Finally, VLM-based instruction annotation enables scale but may introduce linguistic bias or occasional semantic mismatch.

\paragraph{Acknowledgements.}
We gratefully acknowledge Yujun Shen for insightful suggestions and help. We also thank the Realsee Team for providing access to the RealSee3D dataset.

\bibliographystyle{unsrtnat}
\bibliography{references}

\newpage
\appendix

\section{Data Analysis and Statistics}

In this section, we provide detailed statistics and analysis of the datasets used to build the interactive environments in Image2Sim, as well as the distributions of the generated navigation data.

\subsection{Environment Dataset Statistics}
To build interactive 3D environments for Image2Sim, we aggregate diverse real-world and synthetic datasets. 
Table~\ref{tab:dataset_stats} summarizes their key statistics, including scenes, panoramic images, pinhole images, and navigable area. 
Overall, Image2Sim ingests 19,936 scenes with over 617K panoramas, 1.16M pinhole images, and nearly 1.2M m$^2$ of navigable space. 
\textbf{To prevent data leakage, we exclude 29 validation and test scenes from the 90 Matterport3D scenes during navigation training.}

\begin{table}[hbt!]
\centering
\caption{Detailed statistics of the 3D scenes used in Image2Sim.}
\label{tab:dataset_stats}
\resizebox{\textwidth}{!}{%
\begin{tabular}{lrrrr}
\toprule
\textbf{Dataset Name} & \textbf{Scenes} & \textbf{Pano Images} & \textbf{Pinhole Images} & \textbf{Navigable Area (m$^2$)} \\
\midrule
RealSee3D-Real & 1,000 & 24,264 & 0 & 75,198.90 \\
RealSee3D-Synthesis & 9,000 & 274,790 & 0 & 727,493.97 \\
Structured3D & 3,496 & 21,812 & 78,461 & 155,112.76 \\
ScanNet & 1,613 & 0 & 132,418 & 7,289.50 \\
ARKitScenes & 3,345 & 0 & 957,701 & 18,492.32 \\
Matterport3D & 90 & 9,684 & 0 & 31,209.51 \\
HM3D & 900 & 176,917 & 0 & 105,900.34 \\
Gibson & 492 & 109,918 & 0 & 75,418.18 \\
\midrule
\textbf{TOTAL} & \textbf{19,936} & \textbf{617,385} & \textbf{1,168,580} & \textbf{1,196,115.46} \\
\bottomrule
\end{tabular}%
}
\end{table}

\subsection{Generated Navigation Data Statistics}
Based on the constructed scenes, our automated data engine synthesized over 10 million vision-language-action training samples. The generated navigation instructions cover a diverse range of text lengths and styles, effectively mitigating instruction bias. Specifically, the instruction styles are distributed evenly across three categories:
\begin{itemize}[leftmargin=5mm]
    \item \textbf{R2R-style (Path-following):} 3,454,136 samples (33.42\%)
    \item \textbf{REVERIE-style (Object-oriented):} 3,461,321 samples (33.49\%)
    \item \textbf{Demand-style (Human-centric):} 3,419,752 samples (33.09\%)
\end{itemize}

\begin{figure}[hbt!]
    \centering
    \includegraphics[width=0.7\textwidth]{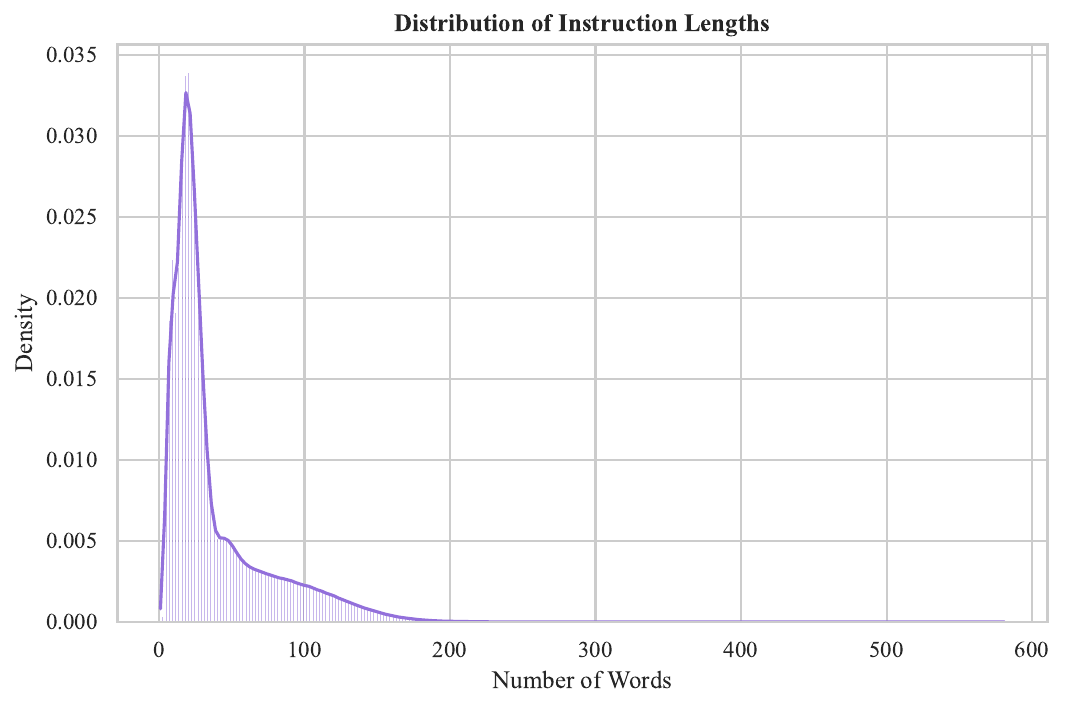}
    \vspace{-10pt}
    \caption{Distribution of instruction lengths (number of words) in the generated dataset.}
    \label{fig:instruction_lengths}
\end{figure}

As illustrated in Figure~\ref{fig:instruction_lengths}, the length of the generated instructions ranges from 1 to 581 words. The distribution exhibits a mean of 38.1 words and a median of 24.0 words, 
which reflects a healthy mix of concise commands and detailed navigational guidance.

\begin{figure}[hbt!]
    \centering
    \includegraphics[width=0.7\textwidth]{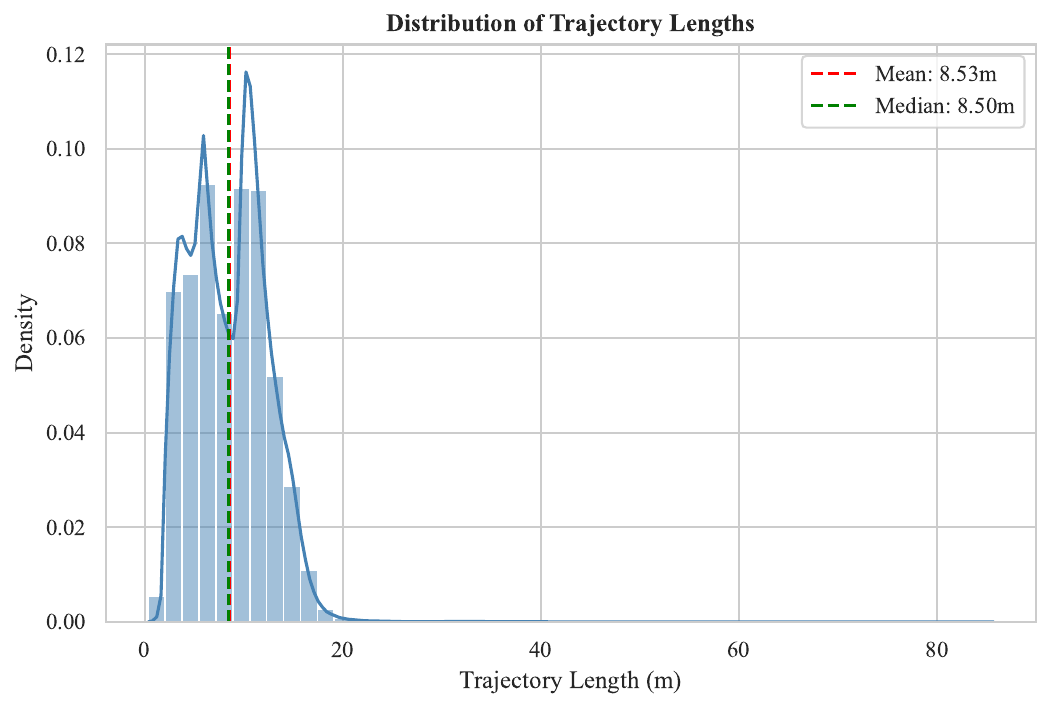}
    \vspace{-10pt}
    \caption{Distribution of trajectory lengths (in meters).}
    \label{fig:trajectory_lengths}
\end{figure}

\begin{figure}[hbt!]
    \centering
    \includegraphics[width=0.7\textwidth]{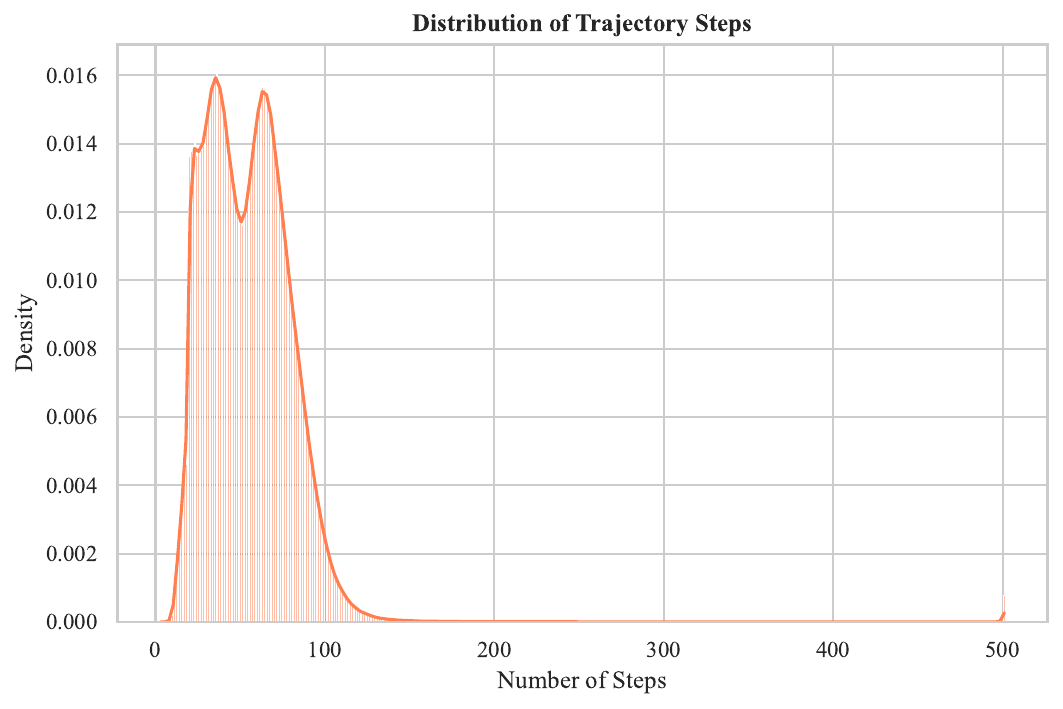}
    \vspace{-10pt}
    \caption{Distribution of trajectory steps (number of discrete actions).}
    \label{fig:trajectory_steps}
\end{figure}

Figure~\ref{fig:trajectory_lengths} and Figure~\ref{fig:trajectory_steps} present the distributions of the generated physical trajectories. The trajectory lengths exhibit a mean of 8.53 meters and a median of 8.50 meters, with a standard deviation of 3.86 meters. The overall travel distances range from a minimum of 0.37 meters to a maximum of 85.72 meters to ensure the navigation policy learns both local target approaches and long-horizon exploration. In terms of agent actions, the trajectories require a mean of 54.7 steps and a median of 53.0 steps to complete, bounded by a maximum of 501 steps.

\section{Clarifications on Experimental Setup and Performance Analysis}
\label{sec:clarifications}

In this section, we provide further details regarding the baseline configurations in our novel-view rendering evaluation, as well as a deeper analysis of the primary source of performance gains in our navigation experiments.

\subsection{Details on Novel-View Rendering Baselines (Table~\ref{tab:combined_results})}
\label{subsec:rendering_baselines}

To ensure a fair and rigorous comparison among rendering baselines with heterogeneous input requirements, we standardized the input pre-processing for the experiments reported in Table~\ref{tab:combined_results}. 

Specifically, the pure generative baselines, DiT360 and SE3DS, take the base images rendered by our Gaussian splatting model as conditional inputs and output native panoramic RGB images. 
In contrast, AnySplat strictly requires pinhole camera inputs. To accommodate this, we perspectively projected the panoramic images into six pinhole RGB-D images, each with a $90^\circ$ field of view. These projected pinhole images were then fed into AnySplat for processing, and the resulting reconstructed 3D Gaussian scene was subsequently rendered using a panoramic Gaussian renderer. To maintain strict experimental alignment, our Image2Sim (Pinhole input) utilizes this exact same panoramic-to-pinhole input processing protocol.

\subsection{Source of Navigation Performance Gains (Table~\ref{tab:comprehensive_vln_results} and Table~\ref{tab:image2nav_scaling})}
\label{subsec:navigation_gains}

We explicitly clarify that the substantial performance improvements demonstrated by our Image2Nav model in Table~\ref{tab:comprehensive_vln_results} are fundamentally attributed to the scale and diversity of the Image2Sim-generated data 
instead of architectural inductive biases or complex navigation modules. 

As evidenced by the scaling analysis in Table~\ref{tab:image2nav_scaling}, when the Image2Nav model is trained exclusively on the standard human-annotated R2R and RxR datasets, its performance is significantly lower than almost all competing baselines listed in Table~\ref{tab:comprehensive_vln_results}. 
Although the Image2Nav architecture provides the critical advantage of high training efficiency, its competitive edge and superior cross-simulator generalization capabilities are entirely driven by the 10M synthetic training samples generated by our pipeline.

\subsection{Hyperparameters and Loss Weights}
\label{subsec:hyperparameters}

The Image2Sim neural simulator is trained using the AdamW optimizer with an initial learning rate of $1\times 10^{-4}$ and a weight decay of $1\times 10^{-4}$. The learning rate is decayed following a Cosine Annealing schedule down to a minimum of $1\times 10^{-6}$ over a total of $400K$ training iterations. The model is trained using Distributed Data Parallel (DDP) with a per-GPU batch size of 2.

We adopt a three-stage curriculum learning strategy to stabilize the one-step rendering and distillation process. The loss weights dynamically adjust across these stages to prioritize different learning objectives:
\begin{itemize}[leftmargin=5mm]
    \item \textbf{Stage 1: Geometry Initialization (0 - 100K iterations).} The network focuses on grounding the 3D representation. The self-distillation is disabled. The perceptual loss weight $\lambda_{perc}$ is set to 1.0.
    \item \textbf{Stage 2: Flow and Distillation (100K - 300K iterations).} The EMA teacher is introduced (decay rate $\gamma = 0.999$) to guide the student. The distillation loss weight $\lambda_{distill}$ is set to 2.0. To enhance visual fidelity, the perceptual loss weight $\lambda_{perc}$ is increased to 2.0, and an adaptive spatial focus weight is applied to the flow loss to emphasize unobserved regions.
    \item \textbf{Stage 3: High-Frequency Refinement (300K - 400K iterations).} The encoder backbone, feature upsampler, semantic aligner, and the deep transformer bottleneck are completely frozen. The network purely optimizes the decoder to sculpt high-frequency pixel details. The perceptual loss weight $\lambda_{perc}$ is further increased to 4.0.
\end{itemize}

For the base loss terms throughout all stages, the reconstruction loss weight $\lambda_{rec}$ and semantic alignment loss weight $\lambda_{align}$ are both set to 0.1. The flow loss $\lambda_{flow}$ incorporates a saturation penalty term with a coefficient of 0.2 to prevent color degradation.

\subsection{Real-World Deployment Details}
\label{subsec:real_world_deployment}

We validate the effectiveness of our Image2Nav model in real-world scenarios using the Hello Robot Stretch 3. The robot is equipped with a head-mounted Intel RealSense D435i RGB-D camera, which captures streaming posed RGB-D images. 

For inference, we deploy the model on a remote workstation equipped with an NVIDIA RTX 4090 GPU and 64GB of RAM. The workstation handles the computationally intensive vision-language navigation reasoning and communicates with the robot via a local area network (LAN) over WiFi. The experiments are conducted in a physical home-like environment comprising a living room, kitchen, meeting room, and office. Crucially, to strictly evaluate zero-shot generalization, none of the objects or scene layouts in this physical environment are included in the training dataset.

\section{Detailed Architecture of Image2Sim}

\begin{figure}[h]
    \centering
    \includegraphics[width=\textwidth]{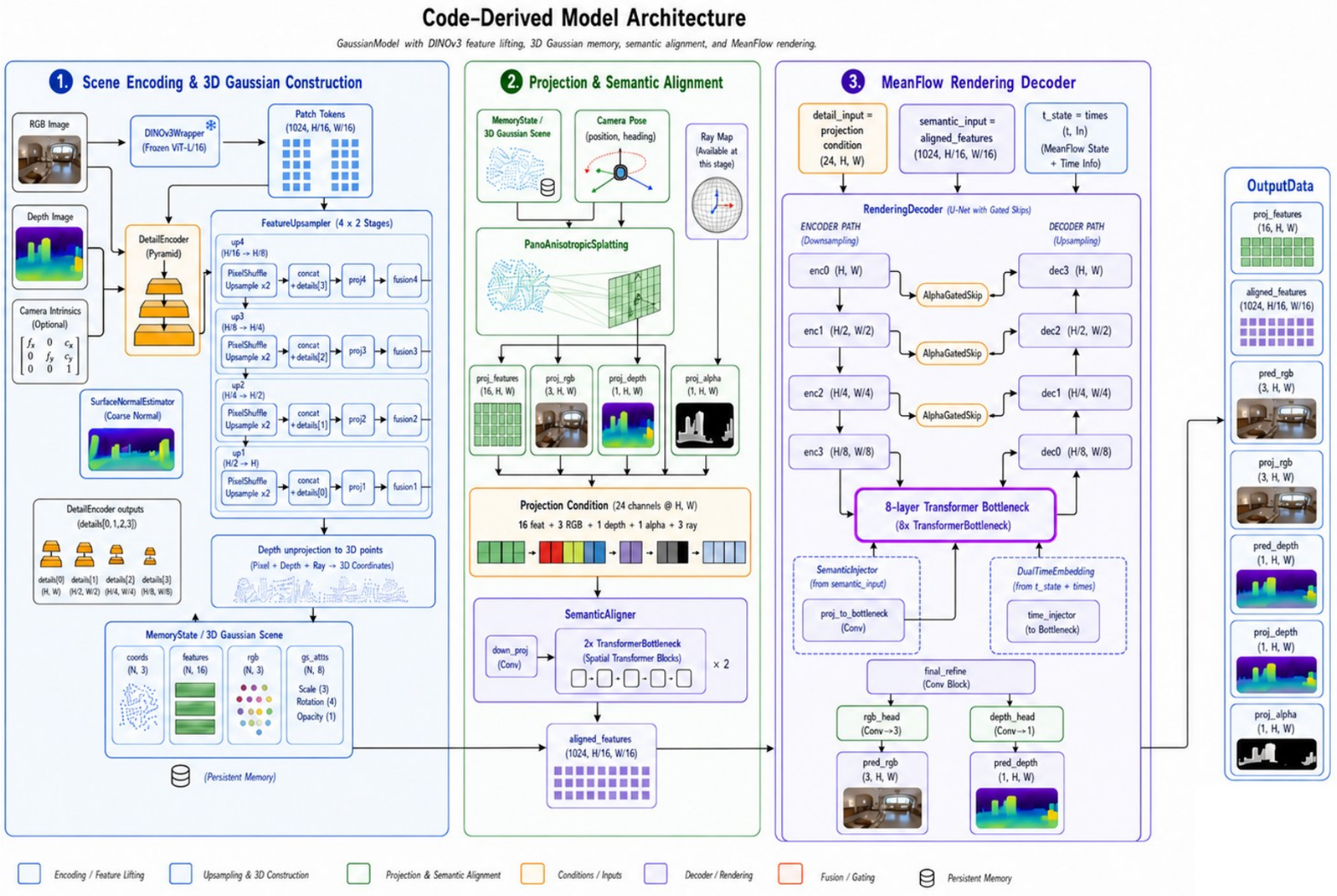}
    \vspace{-25pt}
    \caption{Model Architecture of Image2Sim Renderer (Please zoom in for better viewing).}
    \label{fig:architecture}
\end{figure}

In this section, we provide the comprehensive architectural specifications of Image2Sim, structured to strictly mirror the code implementation. As illustrated in Figure~\ref{fig:architecture}, the framework is inherently modular, explicitly decoupling 3D spatial anchoring from photorealistic generative completion. The architecture operates in three primary stages: Scene Encoding, Projection and Semantic Alignment, and MeanFlow Rendering.

\subsection{Stage 1: Scene Encoding \& 3D Gaussian Construction}

The first stage constructs a persistent 3D Gaussian scene from input visual captures in a feed-forward manner without per-scene optimization. 

\paragraph{Input Representation and Format-Agnostic Ray Directions.}
The initial inputs include RGB images, metric depth, and explicitly computed camera ray directions to ensure the \texttt{FeatureUpsampler} is robust to different camera geometries. The \texttt{get\_ray\_map} function dynamically generates these ray directions based on the input type:
\begin{itemize}[leftmargin=5mm]
    \item \textbf{Panoramic Inputs:} Operates on an equirectangular grid where the pixel coordinates are mapped to longitude $\phi \in [-\pi, \pi]$ and latitude $\theta \in [-\pi/2, \pi/2]$. The resulting unit vectors are computed as $(x, y, z) = (\cos\theta\sin\phi, \cos\theta\cos\phi, \sin\theta)$.
    \item \textbf{Pinhole Inputs:} Utilizes provided intrinsic matrices (focal lengths $f_x, f_y$ and principal points $c_x, c_y$) to compute Z-forward anti-projections: $x = (u - c_x)/f_x, y = (v - c_y)/f_y, z = 1$, which are then $L_2$-normalized into direction vectors.
\end{itemize}

\paragraph{Surface Normal Estimation.}
To provide geometric priors while suppressing high-frequency noise from raw depth sensors, a \texttt{SurfaceNormalEstimator} computes coarse surface normals. The depth map $D$ is first smoothed via a $3 \times 3$ average pooling. Normals are then extracted using $5 \times 5$ Sobel filters ($S_x, S_y$) scaled by $1/48$, which provides a larger receptive field than standard $3 \times 3$ gradients. A constant $Z$-bias of $0.5$ is appended before $L_2$-normalization to suppress lateral random noise.

\paragraph{Dual-Stream Feature Extraction and Fusion.}
The visual and geometric contexts are extracted via a dual-stream design:
\begin{itemize}[leftmargin=5mm]
    \item \textbf{Semantic Stream:} A frozen \texttt{DINOv3Wrapper} (ViT-L/16) extracts rich semantic patch tokens at a resolution $H/16$ with 1024 channels.
    \item \textbf{Detail Stream:} A \texttt{DetailEncoder} processes the 10-channel concatenated input (RGB, depth, normals, ray maps) to build a spatial feature pyramid at $\{H, H/2, H/4, H/8\}$ resolutions.
\end{itemize}
The \texttt{FeatureUpsampler} fuses these streams across four stages. Each stage utilizes a \texttt{PixelShuffleUpsample} operation and fuses features using a \texttt{PanoBlock} (incorporating depth-wise Pano-convolutions and ConvNeXt-V2-style inverted bottlenecks with Global Response Normalization). The network maps these into 16-dimensional scene features and 8-dimensional 3D Gaussian attributes. Rotation quaternions $q$ are factorized into a base rotation $q_{base}$ (aligned to the surface normal) and a learned residual $q_{res}$ to ensure stable initialization. These attributes are unprojected into a persistent \texttt{MemoryState}.

\subsection{Stage 2: Projection \& Semantic Alignment}

Given a target camera pose, the 3D representation is mapped back into the 2D observation space via \texttt{PanoAnisotropicSplatting}, producing a set of raw projections: 16-channel features, 3-channel RGB, 1-channel depth, and a 1-channel alpha (opacity) map.

\paragraph{Semantic Aligner.}
To ensure strict semantic fidelity, a \texttt{SemanticAligner} processes a 24-channel condition tensor (splatted features, RGB, depth, alpha, and target ray map). It applies strided \texttt{PanoConv2d} layers to downsample the spatial resolution to $H/16$ and refines the tokens using two \texttt{TransformerBottleneck} blocks, generating aligned contextual tokens (1024 channels).

\subsection{Stage 3: Pixel Flow Rendering Decoder}

The \texttt{RenderingDecoder} synthesizes the photorealistic panoramic RGB-D observation using a continuous-time MeanFlow formulation. 

\paragraph{Encoder Path and Deep Bottleneck.}
The projection condition is combined with the noisy flow state $z_t$ and downsampled to a resolution of $H/16$. Here, an 8-layer \texttt{TransformerBottleneck} infers global scene structures.
\begin{itemize}[leftmargin=5mm]
    \item \textbf{Semantic Injection:} A \texttt{SemanticInjector} utilizes SPADE-like spatial modulation to infuse 
    DINOv3-aligned tokens into the bottleneck. A simple SiLU and $1 \times 1$ convolution predict scale ($\gamma$) and shift ($\beta$) parameters, applied as $F'_{spatial} = \text{LayerNorm}(F_{spatial}) \odot (1 + \gamma) + \beta$.
    \item \textbf{Temporal Conditioning:} A \texttt{DualTimeEmbedding} module encodes both the absolute integration time $t$ and step size $h = t - r$ using sinusoidal embeddings. These are fused via an MLP to modulate the bottleneck features.
\end{itemize}

\paragraph{Alpha-Gated Skip Connections.}
The decoder upsamples features across four stages. To prevent noisy 3DGS projections from corrupting generated regions, standard U-Net skips are replaced with \texttt{AlphaGatedSkip} modules. An \texttt{AlphaPyramid} generates multi-scale alpha masks (using $2 \times 2$ average pooling) to guide each decoder level. The gating mechanism works as follows:
\begin{itemize}[leftmargin=5mm]
    \item \textbf{Adapter:} Encoder features $F_{enc}$ are aligned to the decoder space using a $1 \times 1$ \texttt{PanoConv2d}, LayerNorm, and GELU activation 
    to get $F'_{enc}$.
    \item \textbf{Mixer:} A gate $g$ is predicted by concatenating the decoder feature $F_{dec}$, adapted encoder feature $F'_{enc}$, and the corresponding multi-scale alpha mask $\alpha$:
    \begin{equation}
        g = \sigma(W_{gate} [F_{dec} \oplus F'_{enc} \oplus \alpha])
    \end{equation}
    \item \textbf{Zero-Initialization Prior:} Crucially, the $1 \times 1$ convolution $W_{gate}$ is zero-initialized with a constant positive bias of $2.0$. 
    This yields an initial gate value of $\sigma(2.0) \approx 0.88$, which encourages the network to strongly trust the 3DGS projection during early training to accelerate convergence. The decoder generation branch $(1-g)$ is only emphasized in regions where the projection is unreliable.
\end{itemize}
The final output of the skip connection is computed as $F_{out} = g \odot F'_{enc} + (1 - g) \odot F_{dec}$.

\paragraph{Prediction Heads.}
The network branches into an RGB Head (Sigmoid activation) and a Depth Head. The Depth Head explicitly uses a ReLU activation instead of Sigmoid 
to accommodate unconstrained metric depth ranges and prevent gradient vanishing in expansive environments.

\section{Details of Motion Simulation and Trajectory Generation}

To support physically grounded embodied learning, Image2Sim incorporates a geometry-aware motion simulation engine. Due to space limitations in the main paper, we detail the complete pipeline for navigable area extraction, graph construction, and kinematic simulation below.

\begin{figure}[h]
    \centering
    \includegraphics[width=\textwidth]{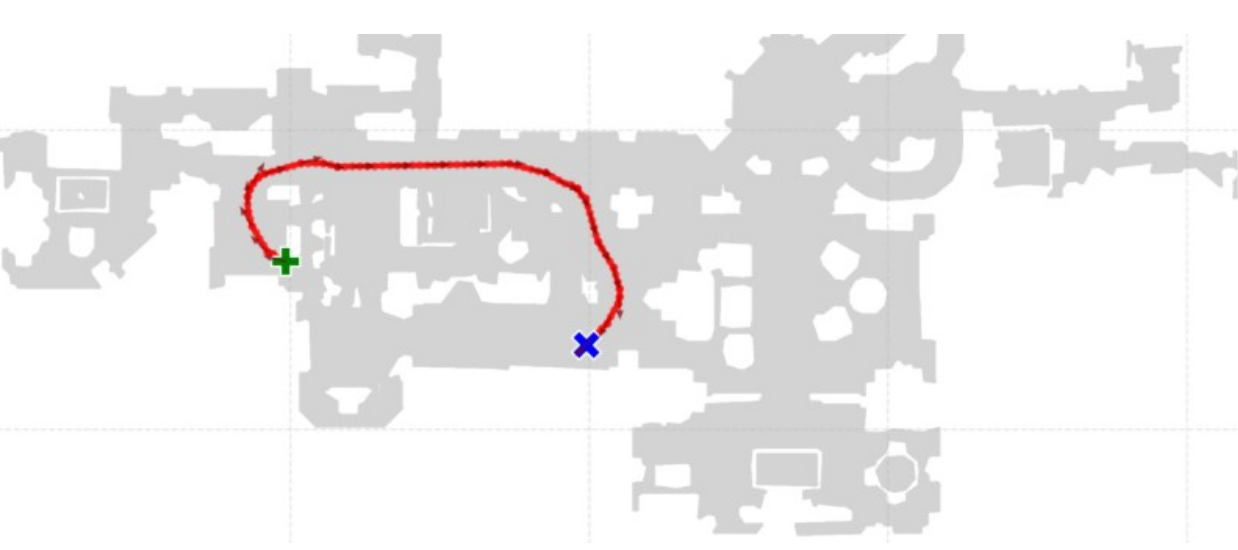}
    \vspace{-15pt}
    \caption{Visualization of navigable voxels and path planning.}
    \label{fig:path_planning}
\end{figure}

\subsection{Navigable Area Extraction and Denoising}
Before path planning, we extract valid navigable regions from the raw RGB-D observations. We employ Mask2Former (Swin-Large) to segment semantic navigable areas (e.g., floors, rugs, sidewalks, and paths) alongside door masks. To ensure geometric validity and eliminate hallucinations, we process the extracted point cloud through a rigorous cleaning pipeline:
\begin{itemize}
    \item \textbf{Geometric Filtering:} The point cloud is downsampled to a $0.05$m voxel size. We estimate surface normals using a $5 \times 5$ Sobel operator and filter out points with a Z-component (vertical axis) smaller than $0.5$, which effectively removes vertical walls that leaked into the semantic mask. 
    \item \textbf{RANSAC Calibration:} We apply RANSAC plane fitting to robustly identify the ground plane and calibrate the scene's coordinate system to ensure the Z-axis strictly aligns with gravity.
    \item \textbf{Obstacle Denoising and Adaptive Filling:} To prevent noisy obstacle points from digging artificial holes in the floor, we apply radius outlier removal. The remaining valid obstacles project a 2D safety buffer. Finally, we project the ground points into a 2D grid ($0.05$m resolution) and apply morphological closing to adaptively fill holes caused by occlusions or depth sensor failures, re-projecting the filled regions back to 3D using the calibrated plane equation.
\end{itemize}

\subsection{Topology and Geometry-Aware Graph Construction}
Using the cleaned navigable point cloud, we construct a dense, traversable voxel connectivity graph $\mathcal{M}=(\mathcal{V},\mathcal{E})$. 
First, the scene is voxelized into a 3D grid with a resolution of $0.05$m. We compute a 2D clearance map using the Euclidean Distance Transform (EDT) to explicitly measure the distance from any navigable voxel to the nearest obstacle. 

To establish locally reachable neighboring states, we perform GPU-parallel ray marching. For each node, a KD-Tree queries candidate neighbors within a $0.3$m search radius. We then cast vectorized rays between candidate pairs, where an edge is retained only if the line-of-sight is entirely free of collisions within the 3D occupancy grid.

The path planning relies on a collision-aware cost formulation instead of merely the shortest path. The traversal cost between node $i$ and node $j$ is defined as:
$$cost_{i \to j} = d_{i,j} + w_{safe} \cdot \left(\max\left(0, \frac{r_{safe} - c_j}{r_{safe}}\right)\right)^2 + p_{collision}$$
where $d_{i,j}$ is the physical distance, $c_j$ is the EDT clearance at node $j$, $r_{safe}$ is a safety margin ($5 \times$ the agent radius), and $w_{safe}$ is the safety penalty weight. The term $p_{collision}$ introduces a soft linear penalty for severe penetration. This design encourages the planner to favor safer corridors and avoid tight corners. The resulting sparse graph is solved efficiently using Dijkstra's algorithm.  \textbf{Specifically, given a 3D starting coordinate and a target destination as inputs, the global planner utilizes this graph to output an optimal collision-free sequence of discrete 3D waypoints as shown in Figure~\ref{fig:path_planning}}.

\subsection{Kinematic Simulation and Closed-Loop Control}
The discrete paths generated by the planner are initially smoothed using a 5-point sliding window average to eliminate the zigzag artifacts inherent to voxel grids. We then utilize a lookahead pure-pursuit controller to convert the path into smooth, executable motion primitives. 

Our simulator assumes a finite-size agent with a radius of $0.15$m and a camera eye height of $1.25$m. The discrete action space consists of \texttt{MOVE\_FORWARD} ($0.25$m step size), \texttt{TURN\_LEFT} ($15^\circ$), \texttt{TURN\_RIGHT} ($15^\circ$), and \texttt{STOP}, matching standard settings of the Habitat agent. \textbf{All these physical and kinematic parameters are fully configurable by the user to accommodate diverse robotic embodiments.} During the physical rollout, the simulator enforces strict collision constraints:
\begin{itemize}[leftmargin=5mm]
    \item \textbf{Geometric Penetration:} At each micro-step, a local KD-tree query checks for obstacles within the agent footprint. Motions exceeding a maximum step height of $0.15$m or violating the radius clearance are rejected.
    \item \textbf{Wall Sliding:} If a direct forward movement is blocked, the simulator computes the center of mass of the obstructing points. Using the cross-product between the heading vector of the agent and the normal of the obstacle, the engine supports physically plausible sliding along obstacle boundaries.
    \item \textbf{Stuck Detection:} We maintain a 4-step sliding window of the state of the agent. If the net spatial displacement is less than $0.1$m and the total yaw rotation is less than $15^\circ$ over 4 steps, the agent is deemed stuck. 
\end{itemize}

\section{Details of Automated Instruction Annotation Engine}

To construct a high-quality, large-scale vision-language-action dataset, we develop a fully automated pipeline comprising global object mining, topologically-aware trajectory sampling, and multi-modal instruction annotation.

\subsection{Global Object Mining and 3D Grounding}
For goal-oriented tasks (e.g., REVERIE and human demands), the agent must locate specific semantic targets. We utilize a pre-trained open-vocabulary detector, YOLO-World, configured with a comprehensive indoor vocabulary. To ensure a broad coverage, the mining process is executed across the navigable area from different viewpoints.

To reliably ground 2D detections into the 3D scene, we apply a filtering mechanism:
\begin{itemize}
    \item \textbf{Adaptive Confidence and Scale:} We relax the confidence threshold to $0.08$ but enforce a minimum area of the bounding box of $300$ pixels to capture small and long-tail objects while rejecting noise.
    \item \textbf{Depth Validation:} Bounding boxes are projected into 3D using the median depth of the central patch and the camera intrinsic matrix. Detections with a depth greater than $10$m are discarded to ensure spatial accuracy.
    \item \textbf{Duplicate Suppression:} A $0.5$m Euclidean distance threshold is applied to merge overlapping detections of the same category.
\end{itemize}

\subsection{Topologically-Aware Trajectory Sampling}
To prevent the navigation policy from overfitting to specific path lengths or dense clusters of a single scene, we implement a robust trajectory sampling strategy over the pre-built connectivity graph.

\begin{itemize}[leftmargin=5mm]
    \item \textbf{Dynamic Scene Capacity:} The maximum number of sampled trajectories per scene is dynamically scaled based on the physical navigable area bounded between 50 and 1000 episodes to prevent topological redundancy in small rooms.
    \item \textbf{Length-Bucketed Distribution:} Target paths are explicitly sampled using a roulette wheel selection across three length buckets: short (3-6m, 20\%), medium (6-10m, 30\%) and long (10-15m, 50\%). This ensures that the model learns both local manipulation approaches and long-horizon exploration.
    \item \textbf{Spatial Diversity:} We employ Furthest Point Sampling against historical start and end points within the scene to maximize global spatial coverage.
    \item \textbf{Long-Tail Goal Targeting:} At the destination, the agent queries the mined object database. Target objects are selected using an Inverse Document Frequency (IDF) weighting scheme, which heavily penalizes frequent categories to aggressively sample rare long-tail objects. Valid targets must fall within a strict Euclidean distance ($1.5$m for manipulable items, $3.0$m for large objects) and pass a geodesic distance check to prevent selecting objects behind walls.
\end{itemize}

Once the start and end points are determined, the optimal path is calculated using our global planner. A physical rollout is then executed using the step function of the simulator. During this rollout, a 4-step sliding window monitors the agent state. If the accumulated spatial displacement is less than $0.1$m and the yaw rotation is less than $15^\circ$, the agent is classified as stuck and the trajectory is truncated.

\subsection{Multi-Modal Instruction Annotation Pipeline}
Directly translating dense kinematic actions (e.g., $0.25$m forward steps) into natural language often yields overly rigid instructions. To bridge this granularity gap, our annotation pipeline formulates the task as an interleaved image-text reasoning process using the Qwen3-VL (32B) model.

\begin{itemize}[leftmargin=5mm]
    \item \textbf{Odometry-Based Semantic Chunking:} We aggregate the high-frequency physical steps into semantic macro-steps based on accumulated odometry. A new macro-step boundary is triggered when the agent accumulates a spatial displacement of $1.0$ meters, a yaw rotation of $45^\circ$, or executes a terminal \texttt{STOP} action. 
    
    \item \textbf{Interleaved Image-Action Prompting:} For the beginning of each macro-step, we extract a $180^\circ$ ultra-wide field-of-view image from rendered panorama. The accumulated kinematic movement of that step is heuristically mapped into a concise text description (e.g., ``Walk forward'', ``Walk forward while turning right'', ``Turn left''). The VLM is then prompted with a strictly alternating sequence of observations and movements: \texttt{[Image 1] $\to$ [Movement 1] $\to$ [Image 2] $\dots \to$ [STOP]}.
    
    \item \textbf{Joint Quality Assurance and Conditioned Generation:} We integrate automated dataset filtering directly into the generation process. Before writing the instruction, the VLM must explicitly output a \texttt{REASONING} block evaluating consecutive frames for fatal simulation errors, such as visual freezing (agent stuck) or unnatural clipping through solid walls. 
    
    \item \textbf{Decoupled Prompt Matrix:} If the trajectory passes inspection, the VLM applies a decoupled prompt matrix. The prompt dynamically combines syntax and tone constraints to route the generation into one of three distinct styles: fine-grained R2R-style (sequential and path-following), REVERIE-style (object-centric and passive voice), or human-centric Demand (casual interaction requests). Strict anti-hallucination rules are enforced, 
    which requires every mentioned object to be visually grounded in the provided frames.
\end{itemize}


\end{document}